\documentclass[11pt]{openqwen2vl}

\usepackage[
    style=numeric-comp,
    doi=true,
    url=false,
    uniquename=false,
    giveninits=true,
    natbib,
    backend=biber]{biblatex}
\usepackage{colortbl}
\usepackage[utf8]{inputenc} 
\usepackage[T1]{fontenc}    
\usepackage{hyperref}
\usepackage{url}            
\usepackage{booktabs}       
\usepackage{amsfonts}       
\usepackage{nicefrac}       
\usepackage{microtype}      
\usepackage{xcolor}
\definecolor{bluelink}{RGB}{0,113,188}
\definecolor{greenlink}{RGB}{0,188,113}
\hypersetup{
    colorlinks=true,%
    citecolor=green!93!black,%
    filecolor=redlink,%
    linkcolor=black,%
    urlcolor=bluelink
}
\usepackage{tabularx}
\usepackage{amsmath}
\usepackage{multirow}
\usepackage{multicol}
\usepackage{array}
\usepackage{caption}
\usepackage{wrapfig}
\usepackage{enumitem}
\usepackage{tikz}
\usepackage{lipsum}
\usepackage{subcaption}
\usepackage{multirow} 
\usepackage{adjustbox}
\usepackage{algorithm}
\usepackage{algpseudocode}
\usepackage{arydshln}

\usepackage{amssymb}
\usepackage{unicode}  
\usepackage[capitalize]{cleveref} 

\usepackage{pgf}
\usepackage{colortbl}

\usepackage{tipa}

\usepackage{tcolorbox}
\usepackage{rotating}
\usepackage[abs]{overpic}
\usepackage{makecell}
\usepackage{longtable}

\usepackage{tocloft}  

\usepackage[english]{babel}
\usepackage{lineno}
\usepackage{csquotes}

\usepackage{fontawesome}
\usepackage{xspace}
\newcommand{\huggingface}{\raisebox{-1.5pt}{\includegraphics[height=1.05em]{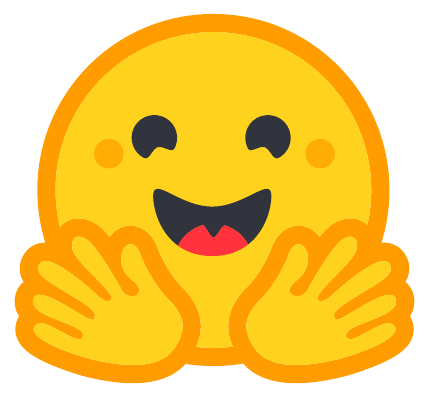}}\xspace}
\newcommand{\github}{\raisebox{-1.5pt}{\includegraphics[height=1.05em]{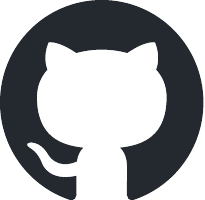}}\xspace}

\newcommand{\worldwideweb}{\raisebox{-1.5pt}{\includegraphics[height=1.05em]{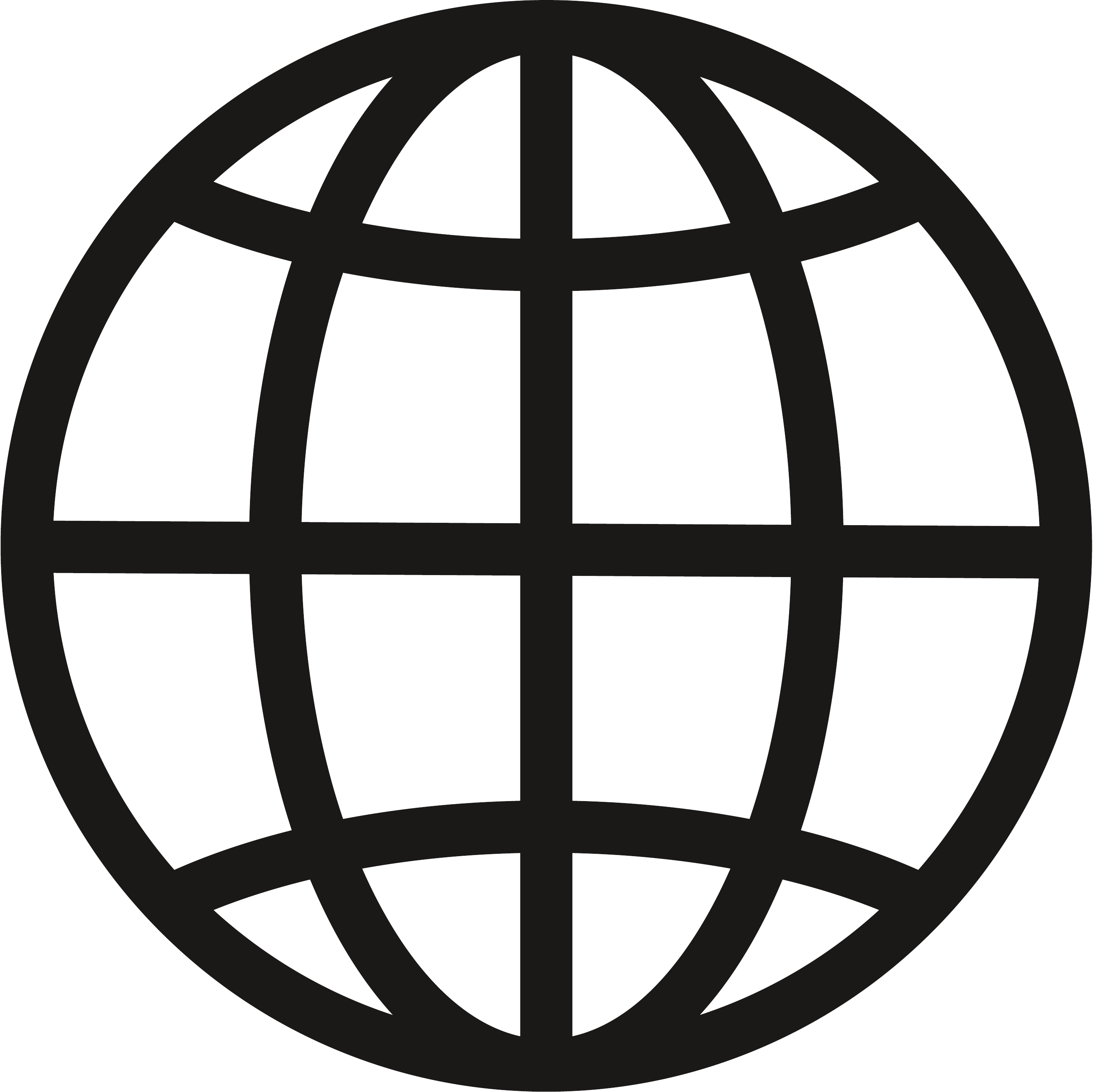}}\xspace}

\definecolor{darkblue}{RGB}{28, 79, 121}
\definecolor{lightgreen}{RGB}{223, 240, 216}
\definecolor{darkgreen}{RGB}{70, 118, 38}
\definecolor{lightblue}{RGB}{217, 237, 247}
\definecolor{highlightgreen}{RGB}{0, 255, 0}

\definecolor{Gray}{gray}{0.93}
\definecolor{uclagold}{rgb}{1.0, 0.7, 0.0}
\definecolor{airforceblue}{rgb}{0.36, 0.54, 0.66}
\definecolor{rosegold}{rgb}{0.72, 0.43, 0.47}
\definecolor{pastelbrown}{rgb}{0.51, 0.41, 0.33}
\definecolor{isabelline}{rgb}{0.96, 0.94, 0.93}
\definecolor{macaroniandcheese}{rgb}{0.98, 0.89, 0.83}
\definecolor{wildblueyonder}{rgb}{0.85, 0.89, 0.95}
\definecolor{mediumtaupe}{rgb}{0.4, 0.3, 0.28}
\definecolor{bluegray}{rgb}{0.4, 0.6, 0.8}
\definecolor{celestialblue}{rgb}{0.29, 0.59, 0.82}
\definecolor{darkorange}{rgb}{1.0, 0.55, 0.0}
\definecolor{cadmiumred}{rgb}{0.89, 0.0, 0.13}
\definecolor{magnolia}{rgb}{0.97, 0.96, 1.0}
\definecolor{pastelblue}{rgb}{0.68, 0.78, 0.81}
\definecolor{persiangreen}{rgb}{0.0, 0.65, 0.58}
\definecolor{steelblue}{rgb}{0.27, 0.51, 0.71}
\definecolor{bluebell}{rgb}{0.64, 0.64, 0.82}
\definecolor{dimgray}{rgb}{0.41, 0.41, 0.41}
\definecolor{splashedwhite}{rgb}{1.0, 0.99, 1.0}
\definecolor{lavendergray}{rgb}{0.77, 0.76, 0.82}
\definecolor{lightgray}{rgb}{0.83, 0.83, 0.83}
\definecolor{lavendermist}{rgb}{0.9, 0.9, 0.98}
\definecolor{lightgreen}{HTML}{f8fcf4}
\definecolor{lightblue}{HTML}{dfebf7}
\definecolor{zeroshot}{rgb}{0.9, 0.9, 0.9}
\definecolor{fourshot}{rgb}{0.8, 0.9, 0.8}
\definecolor{eightshot}{rgb}{0.8, 0.8, 0.9}
\definecolor{sixteenshot}{rgb}{0.9, 0.8, 0.8}

\newcommand\our{\makebox{\text{UniFilter}}}

\title{Train a Unified Multimodal Data Quality Classifier with Synthetic Data}

\renewcommand{\thefootnote}{\fnsymbol{footnote}}

\author{Weizhi Wang$^{1,2}$ \quad Rongmei Lin$^2$ \quad Shiyang Li$^2$ \quad Colin Lockard$^2$ \quad Ritesh Sarkhel$^2$ ~~~~~~~~ Sanket Lokegaonkar$^2$ \quad  Jingbo Shang$^{2,3}$ \quad Xifeng Yan$^1$ \quad Nasser Zalmout$^2$ \quad Xian Li$^2$  \\
$^1$UC Santa Barbara~~~$^2$Amazon Stores Foundational AI~~~$^3$UC San Diego \\
}

\begin{abstract}
The Multimodal Large Language Models (MLLMs) are continually pre-trained on a mixture of image-text caption data and interleaved document data, while the high-quality data filtering towards image-text interleaved document data is under-explored. 
We propose to train an efficient MLLM as a Unified Mulitmodal Data Quality Classifier to Filter both high-quality image-text caption and interleaved data (\our{}).
To address the challenge of collecting diverse labeled multimodal data, we introduce a semi-synthetic approach that leverages readily available raw images and generates corresponding text across four quality levels. This method enables efficient creation of sample-score pairs for both caption and interleaved document data to train \our{}.
We apply \our{} to curate high-quality caption data from DataComp caption dataset and interleaved data from the OBELICS image-text interleaved dataset. MLLMs pre-trained on the filtered data demonstrate significantly enhanced capabilities compared to those trained on baseline-filtered data, achieving stronger zero-shot reasoning and in-context learning capabilities. After visual supervised fine-tuning, these \our{}-induced MLLMs achieve stronger performance on various benchmarks, highlighting the downstream benefits of high-quality multimodal pre-training. We release the synthetic training data used for training UniFilter, the UniFilter model checkpoints, and the high-quality interleaved document subset OBELICS-HQ, curated by UniFilter, to the community for reproduction and further development.
\end{abstract}

\begin{document}
 \maketitle
\setcounter{footnote}{0}  
\renewcommand{\thefootnote}{\arabic{footnote}}  

\begin{center}
    \renewcommand{\arraystretch}{1.2}
    \small
    \scalebox{0.88}{
    \begin{tabular}{rll}
        \worldwideweb & \textbf{Website} & \url{https://victorwz.github.io/UniFilter}\\
        \github & \textbf{Code} & \url{https://github.com/Victorwz/UniFilter}\\
        \huggingface & \textbf{UniFilter-Model} & \url{https://huggingface.co/weizhiwang/UniFilter-Qwen2.5-1.5B} \\
        \huggingface & \textbf{UniFilter-Train-Data} & \url{https://huggingface.co/datasets/weizhiwang/unifilter_train_data} \\
        \huggingface & \textbf{OBELICS-HQ} & \url{https://huggingface.co/datasets/weizhiwang/OBELICS_HQ_5M_UniFilter}\\
    \end{tabular}
    }
\end{center}

\newpage

\section{Introduction}
Large-scale multimodal datasets significantly motivates the recent advances in Vision Language Models (VLMs)~\citep{clip,blip,simvlm} and Multimodal Large Language Models (MLLMs)~\citep{lin2024vila,mckinzie2024mm1,idefics2,blip3}. The scaled up data allows the MLLMs to harvest the knowledge in the training corpora to the greatest extent and promotes the state-of-the-art MLLMs.
The MLLMs are trained on a mixture of image-text caption data and interleaved document data to enhance both zero-shot and few-shot capability. Moreover, with the limited computing resources but the overwhelming number of data mining from CommonCrawl Snapshots, the recent large-scale MLLMs are only trained on a data subset for less than one epoch. Therefore, the data quality became the major bottleneck in training stronger models. In selecting high-quality image-text caption dataset, the representative model-based filter, CLIPScore filter~\citep{laion400m,datacomp} has become the predominant data filtering method. However, CLIPScore can only deal with image-text caption data based on the similarity between a single image and a short text caption. It is completely un-explored on how to select high-quality image-text interleaved data, which contains multiple images and long text paragraphs interleaving in one document.

To address this problem, we propose to train an efficient MLLM as a \textbf{Uni}fied Mulitmodal Data Quality Classifier to \textbf{Filter} both high-quality image-text caption and interleaved data (\textbf{UniFilter}). Adopting an MLLM architecture for the proposed data quality classifier effectively overcomes the limitation of CLIPScore, which can only process single image-text pairs. The proposed UniFilter can process both image-text paired and interleaved data and output a float quality score to indicate the quality of this multimodal data sample. Meanwhile, it outperforms CLIPScore on curating high-quality image-text caption data for enhancing both VLM and MLLM pre-training. Simultaneously, it achieves a high inference throughput of 130 samples/s by leveraging Qwen-2.5-0.5b as LLM backbone, sligtly outperforming the CLIPScore method's 128 samples/s on the same hardware. 

The key to train an effective data quality classifier lies in constructing accurate sample-score pairs~\citep{llama3,penedo2024fineweb}. Human annotations for these pairs are costly and challenging to maintain consistency across different annotators. To address this, we propose a novel semi-synthetic multimodal data generation method by leveraging the proprietary MLLMs. Given that the proprietary MLLMs excel in text generation given multimodal inputs and the raw images are readily available, we sample a diverse set of original images from captioned or interleaved data. We then use proprietary MLLMs to generate the full multimodal data following quality requirements across 4 quality levels (Section~\ref{sec:quality_req}), in which a similar 4 level quality score is also used in FineWeb-Edu-Quality-Classifier~\citep{penedo2024fineweb}. Then the synthetic data can be easily constructed as sample-score pairs, with score labels 0, 1, 2, and 3 corresponding to the defined quality levels in the prompts.

In addition to well-designed synthetic training data construction, we conduct comprehensive ablation studies on the effective and efficient multimodal model architecture of \our{} on the held-out validation synthetic sample-score data. We experiment with 6 combinations of choices of the vision encoder, visual projector, and the LLM backbone for constructing the \our{} architecture. The architecture designs of SigLIP-SO-400M vision encoder~\citep{siglip}, adaptive average pooling projector and the Qwen-2.5-0.5B LLM~\citep{qwen2.5} achieves the best trade-off between data quality classification performance and efficiency.

\begin{figure*}[t] 
\centering 
\includegraphics[width=0.9\textwidth]{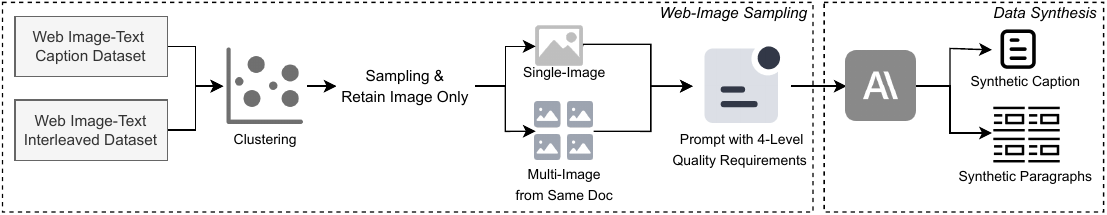}
\caption{The pipeline of semi-synthetic data generation for image-text caption data and interleaved document data.
}
\label{fig:data_pipeline}
\end{figure*}

We conduct comprehensive experiments on image-text caption data and interleaved document data filtering to demonstrate the effectiveness of our method over strong baselines. We firstly validate the priority of \our{} on curating high-quality image-text caption data over strong baselines through the experiments on MLLM pre-training with curated caption data only. Our method \our{} outperforms the state-of-the-art (SOTA) CLIP-based filtering method, Data Filtering Network (DFN)~\citep{dfn} and the SOTAQ MLLM-based data filtering model, MLMFilter~\citep{wang2024finetuned} on all 5 zero-shot VQA datasets. Secondly, pre-training MLLMs on the high-quality image-text interleaved document data curated by \our{} promotes the few-shot learning capabilities of MLLMs by +0.7 and +2.8 average scores on 4-shot and 8-shot VQA performance over the baseline data filters. Finally, instruction-tuned MLLM based on \our{} pre-training outperforms baselines and achieves +3.1 average improvement on VQA tasks and +1.5 improvement on MMMU benchmark, demonstrating the broad benefits of our approach. The summarization of the contributions of \our{} are as follows:
\begin{enumerate}[leftmargin=*]
    \item We introduce \our{}, the first unified approach for filtering both image-text caption and interleaved document data. By leveraging an MLLM-based architecture, UniFilter overcomes the limitations of existing methods that can only process single image-text pairs, enabling effective quality assessment of complex multimodal data structures.
    \item We propose an efficient semi-synthetic data generation method that combines original images with synthetic text across multiple quality levels. This approach addresses the challenge of obtaining diverse, labeled multimodal data for classifier training, enabling scalable and cost-effective creation of high-quality training datasets. 
    \item MLLMs pre-trained on high-quality caption data and interleaved document data filtered by \our{} demonstrate significant performance improvements over models trained with baseline data filtering methods. These gains from high-quality pre-training also persist after the SFT stage, further enhancing model capabilities.
\end{enumerate}

\section{Synthetic Data Construction}
Compared with collecting training data of data quality classification tasks from web-resources or human annotators, the synthetic data can be easily generated on a large scale to provide sufficient training data for models, which is a more feasible solution to empowering the effective training of data quality classifier. Furthermore, by incorporating controlled 4-level quality requirements (Table~\ref{tab:prompt}) into the data generation prompts, synthetic data generation can adhere strictly to these designated quality standards, ensuring clear quality boundaries among data at different quality levels. This proposed approach guarantees data sufficiency of quality classification tasks for effective training of data quality classifier and meanwhile enhances the generalization capabilities of the classifiers on data across multiple quality levels. The synthetic data generation pipeline is shown in Figure~\ref{fig:data_pipeline}.

\subsection{Define Data Quality Requirements}
\label{sec:quality_req}

We firstly design a fine-grained data quality taxonomy on the multimodal data. Instead of using a binary classification of positive and negative, we establish four quality levels: easy negative, medium negative, hard negative, and positive. These four quality levels are designed to capture the spectrum of data quality typically encountered in real-world multimodal datasets. The ``easy negative`` category represents completely irrelevant or nonsensical data, while ``medium negative`` captures data with significant but not entirely unrelated errors. ``Hard negative`` simulates subtle mismatches or minor inaccuracies that are challenging to detect, and ``positive`` represents high-quality, well-aligned multimodal data. 

This granular approach enables our classifier to learn discriminative features across a range of quality levels, enhancing its ability to filter real-world data effectively.
We develop different prompts for both caption and interleaved data to accurately describe each quality level, shown in Table~\ref{tab:prompt}. These prompts guide Claude-3-Sonnet~\citep{claude3} in generating synthetic multimodal data that follows the specified quality requirements. The full prompt giving to Claude-3-Sonnet are presented in Appendix~\ref{appendix:fullprompt}.

\begin{table*}[t]
\centering
\small
\scalebox{1}{
\begin{tabular}{p{65pt}p{360pt}}
\hline    
\toprule
\textbf{Quality Level} & \textbf{Quality Requirements in Prompt} \\
\midrule
Easy Negative & a negative image caption which is completely unrelated to this image. \\
\midrule
Medium Negative & a negative image caption which has remarkable errors in describing the image. \\
\midrule
Hard Negative & a hard negative image caption which has subtle difference with the positive caption. \\
& The negative caption contains only one property error in describing the image. \\
\midrule
Positive & a high-quality, comprehensive, detail-enriched caption for this image. \\
\midrule
\midrule
Easy Negative &  This document should involve many errors in writing and the document itself is not fluent in reading. The images and the text in the document should be completely not related. The images are inserted in inappropriate and arbitrary places in the document. This document should be knowledge limited and has no educational value to be used as textbooks in primary school or grade school teaching. \\
\midrule
Medium Negative & This document is readable but still contains several writing errors. The images and document text are under the same topic and the text contents are still not aligned well to the images. The document is knowledge sparse and has very limited educational value to be used as textbooks in primary school or grade school teaching. \\
\midrule
Hard Negative & This document should involve several errors in writing. The images and the text in the document are partially related. However, the images cannot help the understanding of the text and cannot provide any additional information. The images are inserted in reasonable places in the document. This document should contain several factual or commonsense knowledge errors which makes it inappropriate for educational purposes. \\
\midrule
Positive & This document is a high-quality, comprehensive, detail-enriched document. The images are inserted in the appropriate places in the document to provide additional information to the statement or provide the background information. \\
\bottomrule
\hline
\end{tabular}
}
\caption{Data quality requirements for synthetic caption data and interleaved document data generation.}
\label{tab:prompt}
\end{table*}

\subsection{Semi-Synthetic Data Generation}

The synthetic data on the proposed mutlimodal data quality classification tasks should be diversified and generalized on both image and text sides to train a generalized multimodal data quality classifier. We initially considered generating fully synthetic multimodal data. However, we found that the SOTA image generation models like Midjourney and Dalle-3~\citep{dalle3} are stuck into specific image styles, i.e. carton, due to the post-training adaptations for these diffusion models. Thus, we adopted a semi-synthetic approach: sampling original images from web-crawled caption and interleaved document datasets, while using Claude-3-Sonnet~\citep{claude3} to generate corresponding text at various quality levels.

This semi-synthetic approach offers several advantages. It ensures visual diversity and realism by using real-world images while allowing for controlled text generation at various quality levels. This method is highly scalable and efficient, enabling the creation of large, diverse datasets for classifier training. 

For our semi-synthetic data generation, we selected two prominent datasets as our source data: DataComp for image-text captions and OBELICS for interleaved documents. To enhance the diversity and topic coverage on the image sampling process, we cluster the DataComp-small images into 10k clusters based on their image embeddings extracted by CLIP ViT-L/14. Then we select 4 images from each cluster to get the original 40k image data for the synthetic caption data generation. For the interleaved document data from OBELICS, we compute the average pooling of image embeddings of all images within a single document to create a representative visual embedding for each document. We then clustered the OBELICS dataset into 10k document-level clusters and sampled 40k image-text interleaved documents from these clusters.

Finally, we generate 40k synthetic caption data and 40k synthetic interleaved document data across 4 designed quality levels. We assign the integer quality scores to each synthetic sample while the quality levels of easy negative, medium negative, hard negative, and positive are corresponding to 0,\ 1,\ 2,\ and 3. We held 5\% of 80k data as the validation set for further model developments on the proposed multimodal data quality classification tasks. After collecting original synthetic data, we adopt \texttt{Llama-guard-3-8B}~\citep{llama3} to efficiently scan the synthetic text and ensure there is no safety concerns in the generated texts. We also include 4k non-synthetic high-quality image-caption data from MSCOCO~\citep{mscoco} and Flickr~\citep{flickr} into the final dataset, of which these 4k non-synthetic data are all assigned with quality score of positive. We use the joint sample-score paired data of both image-text caption and interleaved data to train a SINGLE unified multimodal data quality classifier, which can process both image-text caption data and interleaved document data.

\begin{figure*}[t] 
\centering 
\includegraphics[width=0.65\textwidth]{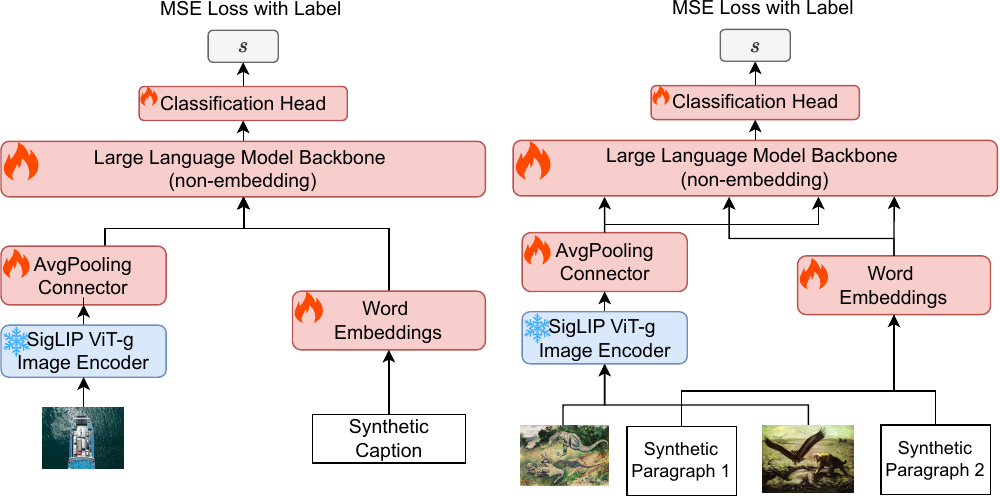} 
\caption{The unified model architecture of \our{} which uses an efficient MLLM to classify the quality scores of both image-text paired data (\textit{Left}) and interleaved data (\textit{Right}).
}
\label{fig:model}
\end{figure*}

\section{UniFilter Architecture}

To achieve a unified architecture to process both the image-text caption and interleaved data, we construct the \our{} based on a MLLM architecture. Figure~\ref{fig:model} presents how a MLLM-based multimodal data quality classifier can process the two major types of image-text data. For the image-text interleaved data, the images and texts are encoded separately with vision encoder and word embedding layer and then reconstructed in original interleaving order. For the caption data, the image encoding and caption embeddings are concatenated and forwarded into the LLM backbone. A trainable one-dimensional classification head is appended on the top of LLM backbone to output a logit indicating the quality score of the input caption or interleaved data sample.


Adopting a MLLM-based data quality classifier can substantially improve the quality classification performance compared with CLIP-based architectures, while the introduced billion-level model parameters will bring huge inference cost to the data quality label inferences on the pre-training data scale. In order to to train the \our{} to be both efficient and capable, we perform comprehensive ablation study on the MLLM architecture of the \our{}. The recent advances on the architecture design of MLLMs~\citep{llava, chen2023internvl,qwenvl} all deploy the modality fusion architecture with 3 major modules of vision encoder, vision-language projector, and the LLM. We inherit this architecture and perform detailed ablations on design choices on different modules. The configuration for model architecture ablations om each module is as follows:
\begin{itemize}[leftmargin=*]
    \item Vision Encoder: We choose two models with different input image resolutions of CLIP-ViT-Large-224px and CLIP-ViT-Large-336px from CLIP model family as well as SigLIP-ViT-SO400m-384px~\citep{siglip} for the ablation studies on vision encoders.
    \item Visual Projector: We consider two type of projector architecture of the non-compressive Multi-Layer Perceptron (MLP) with 2x inner embedding size used in LLaVA~\citep{llava}, and the compressive two-dimensional Adaptive Average Pooling layer with MLP used in DECO~\citep{yao2024deco}.
    \item LLM Backbone: We experiment with four representative small LLMs—Phi-3-mini-3.8B, Gemma-2-2B, and Qwen-2.5 (1.5B and 0.5B), as the base LLM for \our{}. Larger LLMs with more than 4B parameters exceeds efficiency requirements for generating quality scores on pre-training data scale. The pre-trained language modeling head ($\text{Embd}\_\text{Size}\times\text{Vocab}\_\text{Size}$) is deprecated while a newly-initialized classification head ($\text{Embd}\_\text{Size}\times1$) is trained for \our{}. Then the output scalar logit is aligned to the synthetic quality label using Mean-Square-Error (MSE) loss.
\end{itemize}

\begin{table*}[t]
\centering
\small
\scalebox{0.95}{
\begin{tabular}{@{}lllllcc}
\hline    
\toprule
\textbf{LLM}        & \textbf{\begin{tabular}[c]{@{}l@{}}Vision\\ Encoder\end{tabular}} &  \textbf{Projector}   & \textbf{\begin{tabular}[c]{@{}l@{}}\#Tokens\\ \/ per Image\end{tabular}} & \textbf{\begin{tabular}[c]{@{}l@{}}Image\\ Resolution \end{tabular}} & \textbf{\begin{tabular}[c]{@{}c@{}}Validation\\ Acc\end{tabular}}  & \textbf{\begin{tabular}[c]{@{}c@{}}Validation\\ F1\end{tabular}}   \\
\midrule
Phi-3-3.8b & CLIP-L  & MLP & 256  & 224 & 90.8        & 87.1      \\
Phi-3-3.8b & CLIP-L  & AvgPool+MLP & 144 & 336    & 88.7        & 84.0     \\
Phi-3-3.8b & SigLIP-SO-400M  & AvgPool+MLP & 256   & 384    & 91.5  & 87.8       \\
Phi-3-3.8b & SigLIP-SO-400M & AvgPool+MLP & 144  & 384    &91.9  & 88.5     \\
Gemma-2-2b & SigLIP-SO-400M & AvgPool+MLP & 144  & 384    & 88.2        & 83.1        \\
Qwen2.5-1.5b & SigLIP-SO-400M & AvgPool+MLP & 144        & 384    &\bf 95.2       &\bf 94.3     \\
Qwen2.5-0.5b & SigLIP-SO-400M & AvgPool+MLP & 144        & 384    &   94.8  & 93.8 \\
\bottomrule
\hline
\end{tabular}
}
\caption{Ablation studies on the MLLM architecture of \our{}. }
\label{tab:arch}
\end{table*}

\begin{figure}[t] 
\centering 
\includegraphics[width=0.48\textwidth]{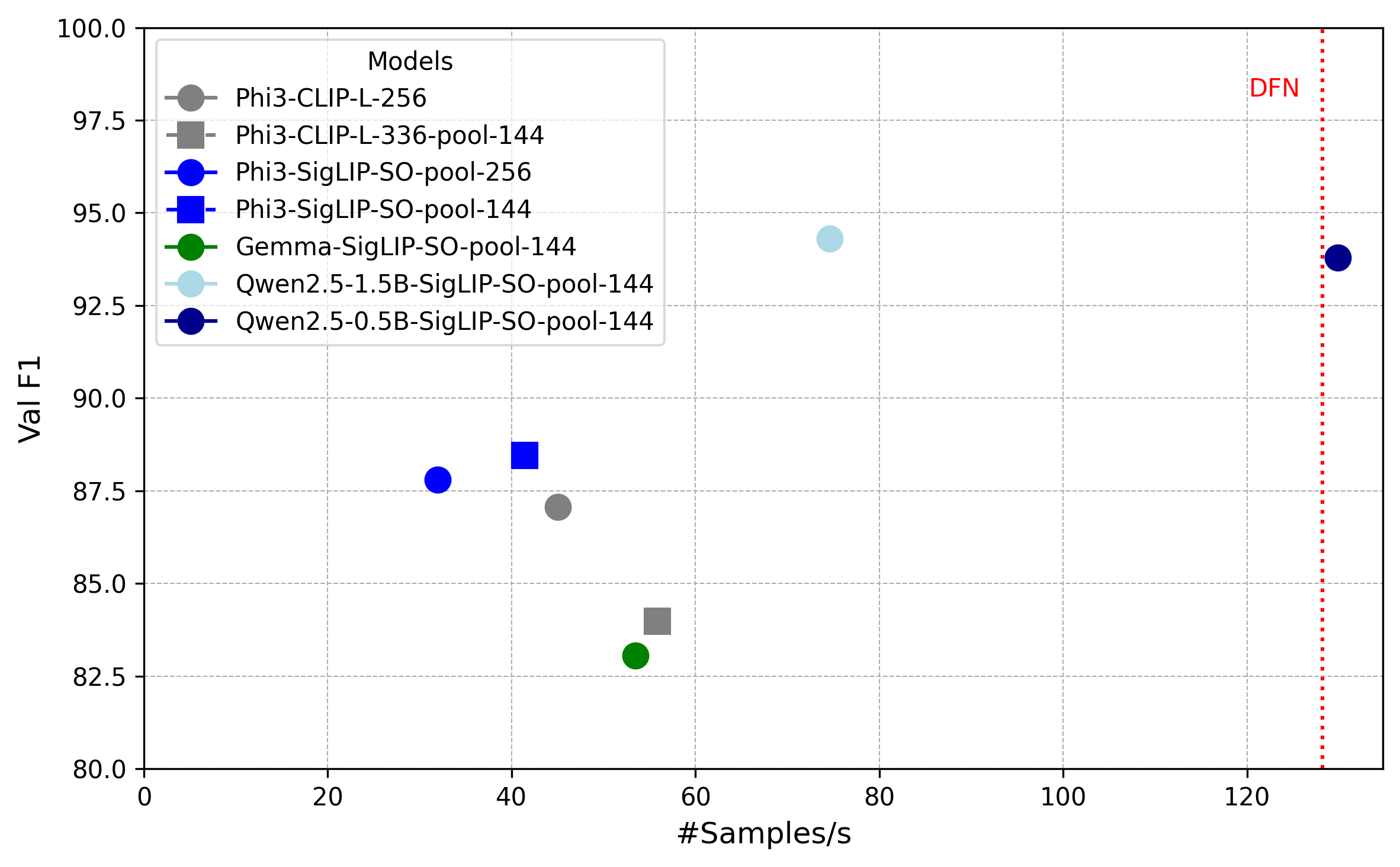}
\caption{The classification F1 versus inference speed of different MLLM architecture ablation configurations on the held validation data of quality classification task.
}
\label{fig:arch}
\end{figure}

Because of the quadratic time complexity of LLMs with respect to number of concatenated multimodal input tokens, the computation efficiency of MLLMs are heavily affected by the number of image tokens for representing one image. Therefore, conducting compression on image patches to fixed number of image tokens is mandatory to ensure the efficiency, especially for high-resolution vision encoders, i.e. SigLIP-so400m-384px and CLIP-Large-336px. \citet{yao2024deco} compares the performance of 4 popular compressive vision projectors, Q-Former~\citep{blip2}, C-Abstractor~\citep{abstractor}, D-Abstractor~\citep{abstractor} and AdaptiveAveragePooling (AvgPool), and the AvgPool significantly outperforms other competitors. Thus, we adopt the two-dimensional AvgPool as the compressive vision projector and compare it with the non-compressive MLP projector. 

We train each \our{} variant for 10 epochs on synthetic contrastive data and the best model is selected based on validation accuracy. The \our{} based on Qwen2.5-1.5b achieves the best quality classification performance while introducing significant computational overhead due to the additional 1b parameters compared with Qwen2.5-0.5b model. Among all group of MLLM architecture configurations, the architecture with SigLIP-SO400M vision encoder, AvgPool visual projector of 144 tokens per image, and the Qwen-2.5-0.5B LLM achieves the best trade-off between quality classification performance and efficiency in Figure~\ref{fig:arch}, which is used as the final \our{} model. Surprisingly, the final \our{} model can achieve comparable inference speed with DFN-CLIP-Large~\citep{dfn}.

\section{Experiments}

\subsection{MLLM Pre-Training on Image-Text Caption Data Only}
\label{sec:mllm-pretrain-caption}


\begin{table*}[t]
\centering
\small
\scalebox{0.95}{
\begin{tabular}{l | cccccc  }
\hline    
\toprule
{\textbf{Methods}} & {\textbf{GQA}} & {\textbf{VQA-v2}} & {\textbf{VizWiz}} & {\textbf{OKVQA}} & {\textbf{TextVQA}}  & {\textbf{Avg.}} \\
\midrule
DFN~\citep{dfn} & 25.8 & 39.6 & 21.6 & 26.0 & 30.7 & 28.7  \\
MLMFilter-Image-Text-Matching~\citep{wang2024finetuned} & 28.3 & 42.7 & 21.7 &	26.0 & 31.6 & 30.2 \\
MLMFilter-Object-Detail-Fulfillment~\citep{wang2024finetuned} & 28.1 & 39.1 & 20.4 & 27.7 & 31.6 & 29.4 \\
MLMFilter-Caption-Text-Quality~\citep{wang2024finetuned} & 28.4 & 40.7 & 20.3 &	27.2 &\bf  35.2 & 30.4 \\
MLMFilter-Semantic-Understanding~\citep{wang2024finetuned} & 24.0 & 40.8 & 18.5 & 23.8 & 29.8 & 27.4  \\
\midrule
\our{} &\bf 29.6 &\bf 43.2 &\bf 22.9 &\bf 28.2 &32.5 &\bf 31.3  \\
\bottomrule
\hline
\end{tabular}
}
\caption{Zero-shot multimodal benchmark results of different pre-trained base MLLMs which are trained on only curated caption data for 5B tokens.}
\label{tab:mllm_pretrain_caption}
\end{table*}

We apply each filtering method, including \our{}, to curate high-quality image-text caption data from the DataComp-medium-128M pool~\citep{datacomp}, and subsequently pre-train separate MLLMs using the datasets curated by the respective filtering methods. The DataComp-medium-128M pool is a noisy, web-crawled image-text caption dataset that employs only basic rule-based filtering, which is specifically designed to evaluate the effectiveness of model-based filtering methods in selecting high-quality caption data.


\textbf{Baselines} We pick the following baseline methods for fair comparisons on MLLM pre-training: 1) Data-Filtering-Network (DFN), a strong CLIP-based data filtering model, which continually pre-trains the OpenAI CLIP-large on high-quality caption dataset for data filtering purpose; 2) MLM-Filter~\citep{wang2024finetuned}, which fine-tunes a MLLM to generate quality scores for caption data filtering with 4 different scoring metrics, Image-Text Matching (ITM), Object Detail Fulfillment (ODF), Caption Text Quality (CTQ), and Semantic Understanding (SU). 

\textbf{Training Setup.} We compare the baseline and our method using the same MLLM architecture and training settings. The model architecture we adopt in MLLM pre-training consists of 3 modules of SigLIP-so400m vision encoder, AvgPool visual projector with 144 tokens per image, and the Phi-3-mini-3.8b LLM. The vision encoder is frozen at all time while other parameters are trainable. To ensure the fair comparisons, we set a fixed \textbf{30\%} fraction of retained high-quality subset from DataComp-medium-128M pool for each filtering method, which can be tokenized into about 6B multimodal tokens for pre-training. Then, each MLLM is trained on filtered image-text caption data by each filtering method for \textbf{5B} multimodal tokens, eliminating the effects of slightly different number of tokens in training for one epoch for each filtered dataset. Other hyper-parameters and details for multimodal pre-training are presented in Appendix~\ref{appendix:details}. Additionally, we perform an ablation study for the filtering fraction hyperparamter in Appendix~\ref{appendix:ablation-fraction}.


\textbf{Evaluation Benchmarks.} We evaluate the zero-shot performance of each \textbf{base} pre-trained MLLMs on 5 visual-question answering datasets, including GQA~\citep{hudson2019gqa}, VQA-v2~\citep{vqav2}, VizWiz~\citep{gurari2018vizwiz}, TextVQA~\citep{singh2019textvqa}, and OKVQA~\citep{okvqa}.
Among the 5 VQA datasets, the VQA-v2 and OKVQA focus on the commonsense knowledge understanding in the images. GQA and VizWiz emphasizes on the scene and spatial understandings and TextVQA lies on evaluating the OCR capability. 

\textbf{Results.} The results in Table~\ref{tab:mllm_pretrain_caption}  demonstrate the superiority of \our{} on curating image-text caption data for enhancing the understanding and reasoning capabilities of base non-sft base MLLMs. Moreover, the base MLLM pre-trained \our{} curated caption data outperforms both DFN and all MLM-Filter metrics on the average performance of 5 multimodal benchmarks, demonstrating the strong generalization and diversity of the \our{}-curated caption data in enhancing the capabilities of MLLMs across OCR, general reasoning, knowledge-reasoning, and scene reasoning. The MLLM trained with UniFilter curated data only lags behind the MLM-CTQ metric on TextVQA task, in which TextVQA dataset requires models to read and reason about text in images. MLMFilter-CTQ metric can effctively differentiate the caption data with great text quality and might be the best performing data filtering metric for OCR or text-rendering related pre-training data.


\subsection{MLLM Pre-Training on Mixed Image-Text Caption and Interleaved Data}
\label{sec:pre-train}

\begin{table*}[t]
\centering
\small
\scalebox{0.95}{
\begin{tabular}{@{}l l l cccccc}
\hline    
\toprule
\multirow{2}{*}{\textbf{Methods}} & \multirow{2}{*}{\begin{tabular}[l]{@{}c@{}}\textbf{\#Train} \\ \textbf{Tokens}\end{tabular}} & \multirow{2}{*}{\textbf{Shots}} & \multirow{2}{*}{\textbf{GQA}} & \multirow{2}{*}{\textbf{VQA-v2}} & \multirow{2}{*}{\textbf{VizWiz}} & \multirow{2}{*}{\textbf{OKVQA}} & \multirow{2}{*}{\textbf{TextVQA}}  & \multirow{2}{*}{\textbf{Avg.}} \\
& & & & & & \\ 
\midrule
\multirow{3}{*}{MM1-3B~\citep{mckinzie2024mm1}} & \multirow{3}{*}{400B} & 0 & & 46.2 & 15.6 & 26.1 & 29.4 & - \\
& & 4 & 
- & 
\cellcolor{lightblue}57.9 &
\cellcolor{lightblue}38.0 & 
\cellcolor{lightblue}48.6 & 
\cellcolor{lightblue}45.3 & -\\
& & 8 & 
- & 
\cellcolor{isabelline}63.6 & 
\cellcolor{isabelline}46.4 & 
\cellcolor{isabelline}48.4 &
\cellcolor{isabelline}44.6 & -\\
\hdashline
\rule{-4pt}{1.05\normalbaselineskip}
\multirow{3}{*}{MM1-7B~\citep{mckinzie2024mm1}} & \multirow{3}{*}{400B} & 0 & & 47.8 & 15.6 & 22.6 & 28.8 & - \\
& & 4 & 
- & 
\cellcolor{lightblue}60.6 & 
\cellcolor{lightblue}37.4 & 
\cellcolor{lightblue}46.6 & 
\cellcolor{lightblue}44.4 & -\\
& & 8 &
- & 
\cellcolor{isabelline}64.6 & 
\cellcolor{isabelline}45.3 & 
\cellcolor{isabelline}51.4 & 
\cellcolor{isabelline}46.3 & -\\
\hdashline
\rule{-4pt}{1.05\normalbaselineskip}
\multirow{3}{*}{BLIP-3~\citep{blip3}} & \multirow{3}{*}{100B} & 0 & - & 43.1 & - & 28.0 & 34.0 &  - \\		
& & 4 & - 
& \cellcolor{lightblue}66.3 & - & 
\cellcolor{lightblue}48.9 & 
\cellcolor{lightblue}54.2 & -\\
& & 8 & 
- & 
\cellcolor{isabelline}66.9 & 
- & 
\cellcolor{isabelline}50.1 & 
\cellcolor{isabelline}55.3 & - \\
\midrule
\multirow{3}{*}{No Filtering} & \multirow{3}{*}{10B} & 0 &  17.6 & 22.5 & 12.2 & 23.9 & 29.1 & 21.1  \\
& & 4 & 
\cellcolor{lightblue}40.4 & 
\cellcolor{lightblue}58.0 & 
\cellcolor{lightblue}37.9 &
\cellcolor{lightblue}44.6 &
\cellcolor{lightblue}38.6 &
\cellcolor{lightblue}43.9 \\
& & 8 & 
\cellcolor{isabelline}40.7 & 
\cellcolor{isabelline}58.6 & 
\cellcolor{isabelline}51.5 & 
\cellcolor{isabelline}45.5 & 
\cellcolor{isabelline}41.0 & 
\cellcolor{isabelline}47.4 \\
\hdashline
\rule{-4pt}{1.05\normalbaselineskip}
\multirow{3}{*}{DFN~\citep{dfn}} & \multirow{3}{*}{10B} & 0 &  21.8 & 36.6 & 16.7 & 20.6 & 30.4 & 25.2 \\
& & 4 & 
\cellcolor{lightblue}40.9	 & 
\cellcolor{lightblue} \bf 60.0	 & 
\cellcolor{lightblue}39.2	& 
\cellcolor{lightblue}43.6 &	
\cellcolor{lightblue} \bf 43.6 &	
\cellcolor{lightblue}45.5 \\
& & 8 & 
\cellcolor{isabelline}41.0 & 
\cellcolor{isabelline}\bf 61.5 & 
\cellcolor{isabelline}45.9 & 
\cellcolor{isabelline}44.9 & 
\cellcolor{isabelline}43.9 & 
\cellcolor{isabelline}47.4 \\
\hdashline
\rule{-4pt}{1.05\normalbaselineskip}
\multirow{3}{*}{\our{}} & \multirow{3}{*}{10B} & 0 & \bf 22.9 & \bf 37.8 &\bf 22.4 &\bf 25.1 &\bf 33.9 &\bf 28.4\\
& & 4 &
\cellcolor{lightblue}\bf 42.2	 & 
\cellcolor{lightblue} \underline{59.7}	 & 
\cellcolor{lightblue}\bf  40.6	& 
\cellcolor{lightblue}\bf 44.8 &	
\cellcolor{lightblue} \underline{43.5} &	
\cellcolor{lightblue}\bf 46.2 \\
& & 8 & 
\cellcolor{isabelline}\bf 42.0 & 
\cellcolor{isabelline}    60.8 & 
\cellcolor{isabelline}\bf 56.3 & 
\cellcolor{isabelline}\bf 46.4 & 
\cellcolor{isabelline}\bf 45.5 & 
\cellcolor{isabelline}\bf 50.2 \\
\bottomrule
\hline
\end{tabular}
}
\caption{Results of MLLMs trained on baseline data and \our{} curated high-quality data for \textbf{10B} tokens. Each 4/8-shot accuracy value is the mean score on 5 random seeds for multimodal in-context learning evaluations. }
\label{tab:mllm}
\end{table*}

Since the experimental results on caption-data pre-training in Section~\ref{sec:mllm-pretrain-caption} have demonstrated the effectiveness and priority of \our{} on caption data filtering, we further investigate the effectiveness of \our{} on filtering interleaved image-text document data to promote the in-context learning capability of MLLM during multimodal pre-training. We use the OBELICS~\citep{laurenccon2024obelics} as the original image-text interleaved document data resource for these experiments.

\textbf{Baseline and Training Setup.} Since there is \textbf{no} effective data filter on filtering high-quality image-text interleaved data on document level, we consider one baseline of no filtering on interleaved data and another DFN
variant baseline for processing interleaved data. For DFN variant, we follow \citet{mmc4} to compute the cosine similarity between each image and each text paragraph in the original interleaved document, and only discard the images of which they do not achieve 0.15 similarity threshold with any of the text paragraph within the same document. MLM-Filter baseline is deprecated here because it can only process the image-text paired caption data and cannot process the interleaved document data. As for our filtering method, the top-15\% high-quality documents, as determined by the \our{} quality scores, are selected as the training data for our MLLM. Given that each induced MLLM will be trained on a mixture of caption and interleaved data, we fix the pre-training caption data as the \our{} curated caption data in Table~\ref{tab:mllm_pretrain_caption} from DataComp-Medium for two baselines and our method to ensure there are no effects from the high-quality caption data side. And then we mix the 5B fixed caption data tokens and 5B image-text interleaved data tokens curated by three methods for each MLLM pre-training. This data mixture ratio of 1:1 is validated in MM1~\citep{mckinzie2024mm1} to be the optimal data mixture ratio to enhance both the multimodal few-shot and zero-shot learning.
We report the 4-shot and 8-shot multimodal in-context learning performance on each VQA dataset for the baselines and our method.
We select 5 random seeds for demonstration example sampling and report the mean score on 5 random seeds as the final task performance.


\textbf{Results.} The results of 0-shot, 4-shot and 8-shot multimodal in-context learning on 5 VQA datasets are presented in Table~\ref{tab:mllm}. We also provide the original results of MM1~\citep{mckinzie2024mm1} and BLIP-3~\citep{blip3} as references even if their training size is 10-40 times larger than ours. Our method \our{} significantly outperforms DFN variant filter baseline on GQA, VizWiz, OKVQA and TextVQA, while slightly lags behind VQA-v2 because the VQA-v2 is constructed from MSCOCO~\citep{mscoco}, which is used as the continue training data of DFN. Finally, the \our{} induced MLLM achieves +0.7 and +2.8 average accuracy improvements over the DFN baseline on 4-shot and 8-shot in-context learning, respectively. The 0-shot in-context learning improvements are much more remarkable than that of 4-shot and 8-shot settings, achieving +3.2 average VQA task improvements. Compared with 4-shot and 8-shot in-context learning with useful instructional information and knowledge from the demonstrations, the 0-shot setting is a more challenging task which relies more on the instruction following capability of models gained from pre-training corpus. Such outstanding 0-shot improvements demonstrate the benefit of effective high-quality interleaved data filtering.

\subsection{Visual Supervised Fine-Tuning}

To further investigate the advantage of high-quality multimodal pre-training on the instruction-tuned MLLM, we perform visual supervised fine-tuning (SFT) on different pre-trained base MLLMs from Section~\ref{sec:pre-train}. The multimodal SFT data is a joint set of visual instruction data from LLaVA-1.5~\citep{llava2} and ShareGPT4V~\citep{chen2023sharegpt4v}. The composition of 575k multimodal SFT data is listed in Appendix~\ref{sec:sft-data-mix}. In addition to 5 VQA datasets,  we include 4 multimodal benchmarks, POPE~\citep{pope}, MMMU (Val)~\citep{mmmu}, MMBench (Dev)~\citep{mmbench}, and MMStar~\citep{mmstar} for comprehensive evaluations towards SFTed models.

\begin{table*}[!t]
\renewcommand\arraystretch{1.05}
\centering
\small
\scalebox{0.83}{
\begin{tabular}{l | cccccc | cccc }
\hline    
\toprule
\multirow{2}{*}{\begin{tabular}[l]{@{}l@{}}\textbf{\ Interleaved} \\ \textbf{Pretrain Data}\end{tabular}} & \multirow{2}{*}{\textbf{GQA}} & \multirow{2}{*}{\textbf{VQA-v2}} & \multirow{2}{*}{\textbf{VizWiz}} & \multirow{2}{*}{\textbf{OKVQA}} & \multirow{2}{*}{\textbf{TextVQA}}  
& \multirow{2}{*}{\begin{tabular}[l]{@{}c@{}}\textbf{VQA} \\ \textbf{Avg.}\end{tabular}}
& \multirow{2}{*}{\textbf{POPE}} & \multirow{2}{*}{\begin{tabular}[l]{@{}c@{}}\textbf{MMMU} \\ \textbf{Val}\end{tabular}}  & \multirow{2}{*}{\begin{tabular}[l]{@{}c@{}}\textbf{MMBench} \\ \textbf{Dev}\end{tabular}} & \multirow{2}{*}{\textbf{MMStar}} \\
& & & & & & & & & \\ 
\midrule
No-Pretrain  & 28.5 & 49.7 & 15.5 & 27.5 & 32.3 & 30.7 & 81.8 & 40.5 & 74.4 & 36.9  \\
\midrule
DFN  & 32.3 & 57.3 & 16.7 & 27.0 & 41.6 & 35.0 & 82.9 & 39.8 & 75.4 & 38.0  \\
\our{} & \bf 33.0 &\bf 60.7 &\bf 19.5 &\bf 32.3 &\bf 44.7 &\bf 38.1 &\bf 83.2 &\bf 42.0 &\bf 77.0 &\bf 38.5 \\
\bottomrule
\hline
\end{tabular}
}
\caption{Zero-shot results of different instruction-tuned MLLMs on VQA datasets and multimodal benchmarks.}
\label{tab:mllm_sft}
\end{table*}

\textbf{Results.} The evaluation results of fine-tuned MLLMs are presented in Table~\ref{tab:mllm_sft}. The fine-tuned MLLM pre-trained on high-quality multimodal data curated by \our{} significantly outperforms the instruction-tuned MLLMs with baseline filtering methods, surpassing the best baseline by +3.1 average VQA accuracy, +1.5 MMMU accuracy, and +1.6 MMBench accuracy. Further, the 0-shot VQA task performance comparisons between the SFT MLLMs and their corresponding base models demonstrate all MLLMs benefit from visual SFT on completing out-of-distribution VQA tasks. The No-Pretrain (SFT-only) baseline significantly lags behind all pre-trained and fine-tuned MLLMs, demonstrating the necessity and benefit of multimodal pre-training. 
\section{Related Work}

\textbf{Data Filtering for LLM and MLLM Pre-training.} 
The family of Phi LLMs~\citep{abdin2024phi} adopt the educational value metric as the data quality metric for filtering high-quality text data for model pre-training, and FineWebEdu-Classifier~\citep{penedo2024fineweb} is an open-source effort on training a data quality classifier for assessing the educational value of web pages. The SOTA open-sourced LLMs, Llama-3 also adopt similar data quality classifier trained on the synthetic sample-score pairs generated by Llama-2-70b, while their classifier training details are not released. In additional, DCLM~\citep{dclm} proposes that instead of training a multi-way quality classifier, a simple binary fasttext~\citep{fasttext} classifier trained on positive instruction tuning data and negative web-crawled data is effective enough to curate high-quality data for SOTA LLM pre-training. In multimodal scenarios, LAION~\citep{laion400m} firstly adopts CLIPScore-based data filtering to select high-quality image-text caption data, and BLIP~\citep{blip} adopts the Cap-Filt data quality boosting method to generate high-quality multimodal training data.

\textbf{Data Quality Classifier Trained with Synthetic Data.} DCLM~\citep{dclm} proposes to construct contrastive data for training a binary text data quality classifier by selecting LLM generated instruction data as positive data and original web-crawled data as negative data. To go beyond the synthetic binary scores, FineWebEdu-Classifier adopts Llama3-70b~\citep{llama3} to generate multi-way quality scores following a well-defined human-drafted score annotation criteria. The data quality classifier to support Llama-3 pre-training also adopts a similar pipeline to instruct Llama-2-chat~\citep{llama2} model to generate the quality scores. In synthetic quality score generation for multimodal data, MLM-Filter~\citep{wang2024finetuned} prompts the GPT-4V to generate the 100-way quality scores on 4 different quality metrics to train a quality classifier for filtering image caption data from 4 distinct perspectives, while AITQE~\citep{aitqe} simlifies MLM-Filter to one unified quality metric on a scale of 0-10.

\section{Conclusion}

We propose an efficient MLLM-based Unified Multimodal Data Quality Classifier to filter both high-quality image-text caption and interleaved data. Pre-training MLLMs on the high-quality data curated by the proposed \our{} can significantly enhance the capability of these general-purpose models on downstream tasks. \our{} overcomes the limitation of being only capable of filtering caption data in CLIP-based data filters and paves a way to steadily improve both zero-shot and few-shot multimodal in-context learning capability of pre-trained and fine-tuned MLLMs via unified multimodal high-quality data filtering.


\printbibliography[heading=bibintoc]

\appendix
\section{Additional Results on DataComp-Medium}
\label{appendix:datacomp-m}

\begin{table*}[t]
\centering
\small
\scalebox{0.83}{
\begin{tabular}{@{}l ll cc c cc}
\hline    
\toprule
\multirow{2}{*}{\textbf{Methods}} & \multirow{2}{*}{\begin{tabular}[l]{@{}c@{}}\textbf{\#Filtered} \\ \textbf{Samples}\end{tabular}} 
& \multirow{2}{*}{\begin{tabular}[l]{@{}c@{}}\textbf{\#Total Training} \\ \textbf{Samples}\end{tabular}} &
\multirow{2}{*}{\begin{tabular}[l]{@{}c@{}}\textbf{ImageNet} \\ \textbf{-1k}\end{tabular}} & \multirow{2}{*}{\begin{tabular}[l]{@{}c@{}}\textbf{ImageNet} \\ \textbf{dist. shifts}\end{tabular}} & \multirow{2}{*}{\textbf{VTAB}} & \multirow{2}{*}{\textbf{Retrieval}} & \multirow{2}{*}{\begin{tabular}[l]{@{}c@{}}\textbf{Avg. over} \\ \textbf{38 datasets}\end{tabular}} \\
    & & & & & &  \\ 
\midrule
No Filtering & 99.6M  & 128M & 16.6   & 14.8       & 25.8  & 21.0   & 25.7     \\
Text-based & 24.8M	& 128M &23.6 & 19.9 & 29.8 & 23.2 & 28.8 \\
Image-based   & 23.7M &  128M  & 23.7 & 19.3       & 29.3  & 23.1 & 28.7     \\
Image-based $\cap$ CLIPScore-30\% & 11.1M  & 128M & 25.5   & 20.8 & 31.2  & 20.0   & 29.7     \\
CLIPScore-30\%~\cite{datacomp}  & 29.6M  & 128M  & 25.5   & 21.6       & 32.5  & 23.1 & 31.5     \\
DFN-Public~\cite{dfn} & 15.0M & 128M & 26.7 & 22.7 & 33.0 & 23.0 & 31.3   \\
DFN-15\%~\cite{dfn} & 15.0M  & 128M  & \textbf{34.4} & \textbf{27.2}    & 35.8  & 25.9 & 34.5     \\
\midrule
\our{}-15\%  & 15.0M  & 128M  & 29.3   & 23.9  & 34.3  & 24.6 & 33.0 \\
\our{}-20\%  & 19.9M  & 128M  & 31.1  & 25.2  & 34.6  & 25.8 & 34.0 \\
\our{}-25\%  & 24.9M  & 128M  & 30.5   & 25.3  & 35.1  & 26.0   & 33.9     \\
\our{}-30\%  & 29.9M  & 128M  & 29.8   & 24.8  & 35.0    & \textbf{27.9} & 34.3     \\
\midrule
\our{}-30\% $\cap$ DFN-15\% & 13.3M  & 128M  & 33.8    & 26.9  & 35.6  & 25.7 & 34.2   \\
\our{}-30\% $\cup$ DFN-15\% & 30.5M  & 128M  & 29.9   & 25.4  & 34.6  & 27.7 & 33.9 \\   
\our{}-25\% $\cup$ DFN-15\% & 26.9M  & 128M  & 31.3   & 25.4  & \textbf{36.3}  & 27.6 & \textbf{35.0} \\    
\bottomrule
\hline
\end{tabular}
}
\caption{Results of \our{} and baselines on DataComp-medium scale caption data filtering benchmark. All the baselines are reproduced for fair comparisons because we can only download 99.6M samples from the original 128M released urls. }
\label{tab:datacomp-m}
\end{table*}

\label{sec:datacomp-m}
The DataComp benchmark~\citep{datacomp} is an image-text data filtering benchmark to systematically compare the performance of different data filtering methods. In DataComp, each data filtering method is required to curate a high-quality subset from a fixed image-text data pool and such selected subset will be used for CLIP VLM pre-training for fair comparisons. The training code and computational budget is controlled across all competing methods to facilitate fair comparison between data filtering methods. The performance of each data filtering method is measured by the evaluation on the zero-shot capabilities of the final induced CLIP model on a suite of 38 classification and retrieval tasks. We select the Medium scale training setting to train ViT-B/32 CLIP models on datasets resulting from baselines and our methods.

\paragraph{Baselines.} We select 5 baselines from original DataComp release, including the No-Filtering, Text-based Filtering, Image-based Filtering, Intersection of Image-based Filtering and CLIPSCore Filtering, and CLIPScore Filtering. We also include the strongest baseline of Data Filtering Network (DFN), which is continually pre-trained based on CLIP to become a stronger data curation model.
The original data pool of DataComp-128M medium scale is released in only image-urls due to copyright considerations and 22.3\% of these image-urls are no longer downloadable at the time of June 2024. To ensure the fair comparisons of different data filtering methods under the same original data pool, we reproduce all baseline methods based on the 99.6M downloadable data. The best filtering fraction of retained high-quality data for CLIPScore and DFN are 30\% and 15\% respectively, which are validated in their original literature~\citep{datacomp,dfn}.

\begin{figure}[ht] 
\centering 
\includegraphics[width=0.6\textwidth]{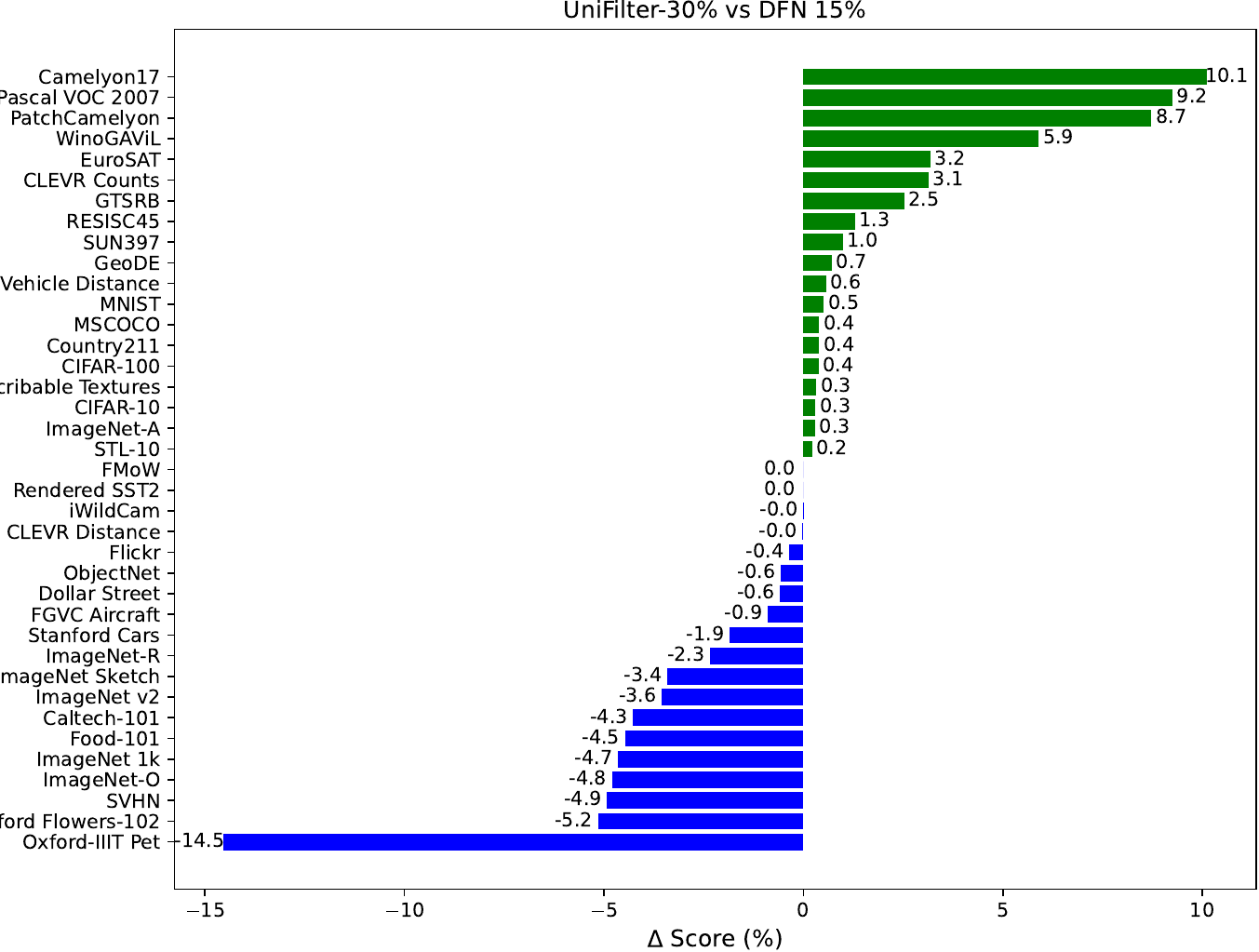} 
\caption{The comparisons between \our{} and DFN on each dataset in the 38 dataset evaluation suite of DataComp Meidum. The \textbf{Green} bars mean \our{} beats DFN.
}
\label{fig:comparison}
\end{figure}

\paragraph{Results.} The results on DataComp-Medium image-text caption data filtering benchmark are shown in Table~\ref{tab:datacomp-m}. We consider 4 different fraction of original data retained as high-quality data, 15\%, 20\%, 25\%, and 30\%. When working individually as the data filter, the \our{} retaining top-30\% of 99.6M data performs best on the average performance over 38 datasets, which is comparable with DFN. We also explore the intersection or union of the filtered high-quality data between ours and DFN's. The union of \our{}-25\% and DFN data achieves the SOTA performance on 38 dataset average score and VTAB average score~\citep{vtab}. Meanwhile, we find that the high-quality data filtered by \our{} can significantly boost the CLIP's performance on image-to-text and text-to-image retrieval tasks. \our{}-30\% achieves SOTA performance on retrieval task average performance and incorporating \our{} data with DFN's data improves DFN's retrieval performance by +1.8 average scores. Such significant improvements on retrieval tasks are also observed in other MLLM-based data filters~\citep{wang2024finetuned}. We suppose that MLLM-based data filter may favor the image captions with high text quality like linguistic fluency and readability, which contributes more to the retrieval tasks than classification tasks.

We also provide the detailed comparisons between DFN and \our{} on each dataset of the 38 dataset suite in Figure~\ref{fig:comparison}. \our{} surpasses DFN on 19 datasets while does worse on ImageNet related datasets because DFN is continually fine-tuned on ImageNet-1k dataset for distribution alignment. We find that \our{} significantly lags behind DFN on Oxford-Pet dataset~\citep{parkhi2012cats} (37.99\% versus 52.52\%). The Oxford-Pet dataset includes fine-grained pet categories (e.g., Abyssinian, American Bulldog) that are rarely represented in publicly available image-text datasets. We also check the performance of other strong baselines like CLIPScore and MLM-Filter on this Oxford-Pet dataset, which are 35.25\% and 38.62\% respectively. We hypothesize that DFN's superior performance on this dataset may be attributed to the extensive coverage of fine-grained pet class names in Apple's internal HQ-357M dataset used for training DFN.

\section{Training Settings of MLLM Pre-Training}
\label{appendix:details}
The training details and hyperparameters for MLLM pre-training are presented in Tab.~\ref{tab:mllmsetting}. We do not follow Qwen-VL~\citep{qwenvl} to perform a separate stage to train the visual projector only. The MLLM pre-training only involves one single stage to update the parameters of visual projector and LLM backbone, while the vision encoder is frozen all the time. To accelerate the MLLM training and avoid too many padding tokens, we perform sequence packing to regroup image-text data at varied length into a fixed context size sequences. A special <|endofchunk|> token is added before the start of every image in an image-text interleaved document to indicate the end of a text paragraph. The MLLM training on 10B mixed multimodal tokens is conducted on 4 A100-40G gpus nodes, and each node contains 8 A100-40G gpus. The training for 10B tokens takes about 640 A100-40G gpu hours. 

\begin{table}[!ht]
\begin{center}
\small
\begin{tabular}{l|c}
\hline
\toprule
\textbf{Details}  & MLLM Pre-Training \\ 
\midrule
Vision Encoder & SigLIP-so400m-384px \\
Visual Projector & 2d Adaptive Average Pooling \\
LLM Backbone & Phi-3-mini-4k-instruct \\
Context Length & 4096 \\
\midrule
Precision & \texttt{BF16}  \\ 
Global Batch Size & 256 \\ 
\# Training Steps & 9537 \\
\# GPUs & 32 A100 \\
Peak LR & 3e-5 \\
\# Warmup Steps Ratio & 3\% \\
LR Scheduler & Cosine LR Decay \\
Weight Decay & 0.01 \\
Adam $(\beta_1, \beta_2)$ & (0.9, 0.98) \\ 
\bottomrule
\hline
\end{tabular}
\caption{Training details for MLLM pre-training on 10B multimodal tokens.}
\label{tab:mllmsetting}
\end{center}
\end{table}


\section{Statistics of Multimodal Datasets}

We list the statistics of the large-scale image-text caption dataset and image-text interleaved document dataset in Tab.~\ref{tab:pretrain-data-stats} as well as their licenses. The filtered high-quality data subset by \our{} will also inherit the original licenses of these datasets and ensure the proper usage of them. All images in two datasets are released in image-urls rather than files, leading to a large-scale invalid data samples. We discard the whole document from OBELICS if any one of the image in the document is invalid. As of June 2024, only a half of OBELICS interleaved documents are fully downloadable. 

\begin{table}[!ht]
\centering
\small
\scalebox{0.85}{
\begin{tabular}{llll}
\hline
\toprule
\bf \multirow{2}{*}{Dataset} &\bf \multirow{2}{*}{\#Samples} & \multirow{2}{*}{\begin{tabular}[l]{@{}l@{}}\textbf{Downloadable} \\ \textbf{\#Samples}\end{tabular}}  &\bf \multirow{2}{*}{License} \\
& & & \\
\midrule
DataComp-Medium & 128M & 99.6M & MIT \\
OBELICS & 141M & 70.5M & CC-BY-4.0 \\
\bottomrule
\hline
\end{tabular}
}
\caption{Pre-Training Multimodal Dataset Statistics.}
\label{tab:pretrain-data-stats}
\end{table}

\section{Statistics of the Curated Interleaved Document Dataset}

\begin{table}[!ht]
\centering
\small
\scalebox{0.9}{
    \begin{tabular}{lcccc}
    \hline
    \toprule
    \multirow{2}{*}{\bf Filter}  & \multirow{2}{*}{\begin{tabular}[l]{@{}c@{}}\textbf{Avg. } \\ \textbf{\#Img.}\end{tabular}}
    & \multirow{2}{*}{\begin{tabular}[l]{@{}c@{}}\textbf{Avg. Text} \\ \textbf{ Len.}\end{tabular}} & \multirow{2}{*}{\begin{tabular}[l]{@{}c@{}}\textbf{Avg. Document} \\ \textbf{Len.}\end{tabular}} & \multirow{2}{*}{\begin{tabular}[l]{@{}c@{}}\textbf{Filtering} \\ \textbf{ Fraction}\end{tabular}} \\
    & & & & \\
    \midrule
    None  & 1.98  & 842.5  & 1125.8   & 100\%   \\
    DFN & 1.88  & 841.1   & 1110.4    & 90.5\%  \\
    UniFilter   & 3.15    & 1627.8 & 2078.3   & 15\%   \\
    \bottomrule
    \hline
\end{tabular}
}
\caption{The curated interleaved document data statistics using different filtering methods.}
\label{tab:data_stats}
\end{table}

We analyze the features and statistics of the curated interleaved document data using different filtering methods in Table~\ref{tab:data_stats}. Compared with the No-Filter and DFN-Filter baselines, the selected high-quality document data contains more images and more text tokens in one document. Additionally, the DFN-CLIP-Filter can only remove the irrelevant images in a document based on cosine similarity, leading to an uncontrollable filtering fraction. The proposed \our{} can achieve document-level filtering and flexibly select the filtering fraction hyperparameter based on the data quality needs.

\section{Visual SFT Data Composition}
\label{sec:sft-data-mix}
We select a comprehensive and diverse task sets for constructing the visual SFT instruction dataset. Since the 5 VQA datasets are used as evaluation benchmarks, the visual instruction data constructed from these 5 VQA dataset are excluded in the joint SFT dataset. For visual instructions, we select LLaVA-Conversations~\citep{llava}, LLaVA-Reasoning~\citep{llava}, ShareGPT4V-Caption~\citep{chen2023sharegpt4v}, OCRVQA~\citep{ocrvqa}, A-OKVQA~\citep{aokvqa}, TextCaps~\citep{textcaps}, RefCOCO~\citep{refcoco}, and VG~\citep{vg}. We use ShareGPT as the text instruction data. The final joint SFT dataset consists of 575k multimodal instructions.

\begin{table}[!ht]
\centering
\small
\scalebox{0.85}{
\begin{tabular}{ll p{53mm}}
\hline
\toprule
Data & Size & Response formatting prompts \\
\midrule
\multicolumn{3}{c}{\textit{Visual Instructions 517k}}    \\
\midrule
Conversation  & 58K & -- \\
Reasoning & 77k & -- \\
ShareGPT4V & 100k & -- \\
\midrule
OCRVQA & 80k & Answer the question using a single word or phrase.\\
\midrule
A-OKVQA & 66K & Answer with the option's letter from the given choices directly. \\
\midrule
TextCaps & 22K & Provide a one-sentence caption for the provided image. \\
\midrule
RefCOCO & 48K & \emph{Note: randomly choose between the two formats} \\
 & & Provide a short description for this region. \\
\cmidrule{1-2}
VG & 86K & Provide the bounding box coordinate of the region this sentence describes. \\
\midrule
\multicolumn{3}{c}{\textit{Text Instructions 40k}}    \\
\midrule
ShareGPT  & 40K & -- \\
\midrule
\multicolumn{3}{c}{\textit{Total 575k}}    \\
\bottomrule
\hline
\end{tabular}
}
\caption{Visual and Text SFT Data Composition.}
\label{tab:data_mixture}
\end{table}

\section{Full Synthetic Data Generation Prompts}
\label{appendix:fullprompt}

The full prompts for both caption data and interleaved data generation are listed in Tab.~\ref{tab:fullprompt}. The "\{multi-level quality requirements\}" are placeholders for integrating the defined quality requirements in Tab.~\ref{tab:fullprompt} into the synthetic data generation prompt.

\begin{table*}[!ht]
\centering
\small
\scalebox{0.95}{
\begin{tabular}{p{400pt}}
\hline    
\toprule
\textbf{Caption Data Generation Prompt} \\
\midrule
You are a helpful assistant to help users write two opposite image captions for the given image in JSON format. The JSON object must contain the following keys:\\
- "topic": a string, a topic word of this image\\
- "positive\_caption": a string, a high-quality, comprehensive, detail-enriched caption for this image. \\
- "negative\_caption": a string, \{multi-level quality requirements\} \\
\\
Please adhere to the following guidelines:\\
- Both captions should be at least \{num\_words\} words long.\\
- Both captions should be in English.\\
- Please avoid using complex or advanced words in the captions. Ensure that the language is suitable for a high school level audience or lower.\\
\\
Your output must always be a JSON object only, do not explain yourself or output anything else. Be creative!\\
\midrule
\rule{-4pt}{1.05\normalbaselineskip}
\textbf{Interleaved Data Generation Prompt} \\
\midrule
You are an assistant to help users to write a document given several images. These images are extracted from a paper, report, or article in which these images are inserted. \\ \\
<guideline>
Please firstly generate a xml tag for each image in order for future generation. For each image, please generate a xml tag like "<img>image description</img>". You need to replace the image description with your generated short description of this image which is less than 5 words.
\\ \\
For the second task, \{multi-level quality requirements\}
\\ \\
Please adhere to the following guidelines when writing this document:\\
- The paragraphs in the document should be in varied length.\\
- The document should contain at least 500 words.\\
- You NEED to use xml tag as the placeholder to indicate the place where an image is inserted into.\\
- You NEED to ensure that all given images are used and considered.\\
- You MUST NOT use the image xml tag within your sentences. You should add them between sentences and paragraphs.\\
- You MUST use each image for ONLY ONCE in the document.\\
\\
Your output must always be a JSON object only. The JSON object must contain the keys of "image\_tags" and "document".\\
\\
</guideline>\\
\\
Now, it is your turn. Please strictly follow the above guidelines in <guideline> xml tags when writing the document.\\
\bottomrule
\hline
\end{tabular}
}
\caption{Prompting templates for synthetic caption data and interleaved document data generation.}
\label{tab:fullprompt}
\end{table*}

\begin{table*}[!ht]
\centering
\small
\scalebox{0.76}{
\begin{tabular}{l c}
\hline    
\toprule
\textbf{Prompt Templates for Demonstration Examples} & \textbf{GQA}  \\
\midrule
<|user|><image>\textbackslash{}n\{Question\}Answer the question with single word or phrase.<|assistant|>\{Answer\}<|end|> & 35.79 \\
\midrule
<bos><|user|><image>\textbackslash{}n\{Question\}Answer the question with single word or phrase.<|assistant|>\{Answer\}<|end|> & 38.46 \\
\midrule
<bos><|user|><image>\textbackslash{}n\{Question\}Answer the question with single word or phrase.<|assistant|>\{Answer\}<|endofchunk|> & 39.64 \\
\midrule
<bos><image>\textbackslash{}n\{Question\}Answer the question with single word or phrase.\textbackslash{}n\{Answer\}<|endofchunk|> & 39.53 \\
\midrule
<bos><image>\textbackslash{}n\{Question\}Answer the question with single word or phrase.\textbackslash{}n\{Answer\}<|endoftext|> & 34.96 \\
\midrule
<bos><image>\textbackslash{}n\{Question\}Answer the question with single word or phrase.\textbackslash{}n\{Answer\}<|end|> & 39.09 \\
\midrule
<bos><image>\textbackslash{}nQuestion: \{Question\}Answer the question with single word or phrase.\textbackslash{}nAnswer: \{Answer\}<|endofchunk|> & \textbf{42.2} \\
\midrule
<bos><image>\textbackslash{}nQuestion: \{Question\}\textbackslash{}nAnswer: \{Answer\}<|endofchunk|> & 37.67 \\
\bottomrule
\hline
\end{tabular}
}
\caption{Ablation studies on the effects of different in-context learning prompt construction templates on the 4-shot performance of GQA using the no-filtering baseline model.}
\label{tab:template}
\end{table*}

\section{Ablations on In-Context Learning Prompt Templates}
\label{appendix:prompt}

The demonstration prompt template affects the performance of multimodal in-context learning. We perform an ablation study on different prompt templates for constructing demonstration examples, shown in Tab.~\ref{tab:template}. The results present that the <|endofchunk|> token is significant to multimodal in-context learning capability of MLLMs. The <|endofchunk|> token is inserted in the end of each text paragraph of the interleaved document data during the data pre-processing. Thus, adding this token to each demonstration example template constructs the few-shot demonstrations into an interleaved document, which may help trigger the model's parametric memory towards the pre-trained knowledge in the interleaved document data.

\section{Ablation Study on Filtering Fraction for Caption Data}

We further investigate the effects of different fraction of retained high-quality subset from the original pool to the performance of pre-trained MLLMs. We perform ablation studies on DFN baseline and \our{} on two filtering fractions of 15\% and 30\%. The results in Table~\ref{tab:ablation-fraction} demonstrates that for both methods 30\% filtering fraction is a better hyperparameter choice compared with retraining only 15\% data. Generally, retraining less data will hurt the data diversity and distribution of curated dataset, and 30\% achieves the best trade-off between highest average quality and data diversity. 

\label{appendix:ablation-fraction}
\begin{table*}[!ht]
\centering
\small
\scalebox{1}{
\begin{tabular}{ll | cccccc  }
\hline    
\toprule
{\textbf{Methods}} & \textbf{Fraction} & {\textbf{GQA}} & {\textbf{VQA-v2}} & {\textbf{VizWiz}} & {\textbf{OKVQA}} & {\textbf{TextVQA}}  & {\textbf{Avg.}} \\
\midrule
DFN & 15\% & 25.7 & 35.8 & 21.6 & 24.9 & 30.3 & 27.7 \\
DFN & 30\% & 25.9 & 41.0 & 21.2 & 24.1 & 29.9 & 28.4 \\
\midrule
\our{} & 15\% & 26.7 & 41.8 & 20.1 & 24.5 & 28.3 & 29.3 \\ 
\our{} & 30\% & \bf 29.6 &\bf 43.2 &\bf 22.9 &\bf 28.2 &\bf 32.5 &\bf 31.3  \\
\bottomrule
\hline
\end{tabular}
}
\caption{Ablation studies on the fraction of retained high-quality subset from the original 128M data pool for MLLM pre-training.}
\label{tab:ablation-fraction}
\end{table*}

\section{Effects of Introducing System Prompts in Multimodal In-Context Learning}

\begin{table}[!ht]
\centering
\scalebox{0.85}{
\begin{tabular}{lcc}
\hline    
\toprule
\bf System Prompts & \bf GQA  \\
\midrule
Phi-3 Default & 40.57 \\
\midrule
LLaVA-1.5 Default & 40.77 \\
\midrule
Claude-3 Generated & 41.60  \\
\bottomrule
\hline
\end{tabular}
}
\caption{Ablation studies on the effects of different system prompt construction in templates on the 4-shot performance of GQA using the no-filtering baseline model. }
\label{tab:system}
\end{table}

In addition to the demonstration prompt template, we also investigate the effects of introducing the system prompts in the in-context learning templates. The results on these ablation studies are presented in Tab.~\ref{tab:system}. We consider 3 different system prompts as follows:
\begin{itemize}
    \item \textbf{Phi-3 Default}: <|system|>You are a helpful assistant.<|end|>
    \item \textbf{LLaVA-1.5 Default}: A chat between a curious human and an artificial intelligence assistant. The assistant gives helpful, detailed, and polite answers to the human's questions.
    \item \textbf{Claude-3 Generated}: You are tasked with answering open-ended questions based on images provided from the visual question answering dataset. Each question may require understanding the visual content of the image, interpreting natural language, and applying commonsense knowledge. Your goal is to generate the most accurate answer based on the image, considering multiple possible interpretations.
\end{itemize}

Concluding from Tab.~\ref{tab:system}, introducing the system prompt in the in-context learning templates promotes the GQA performance of induced MLLM, demonstrating the success of multimodal pre-training to train the base MLLM to follow instructions.

\section{Examples of Claude-3 Generated Contrastive Interleaved Documents.}
\begin{figure*}[!ht] 
\centering 
\includegraphics[width=0.98\textwidth]{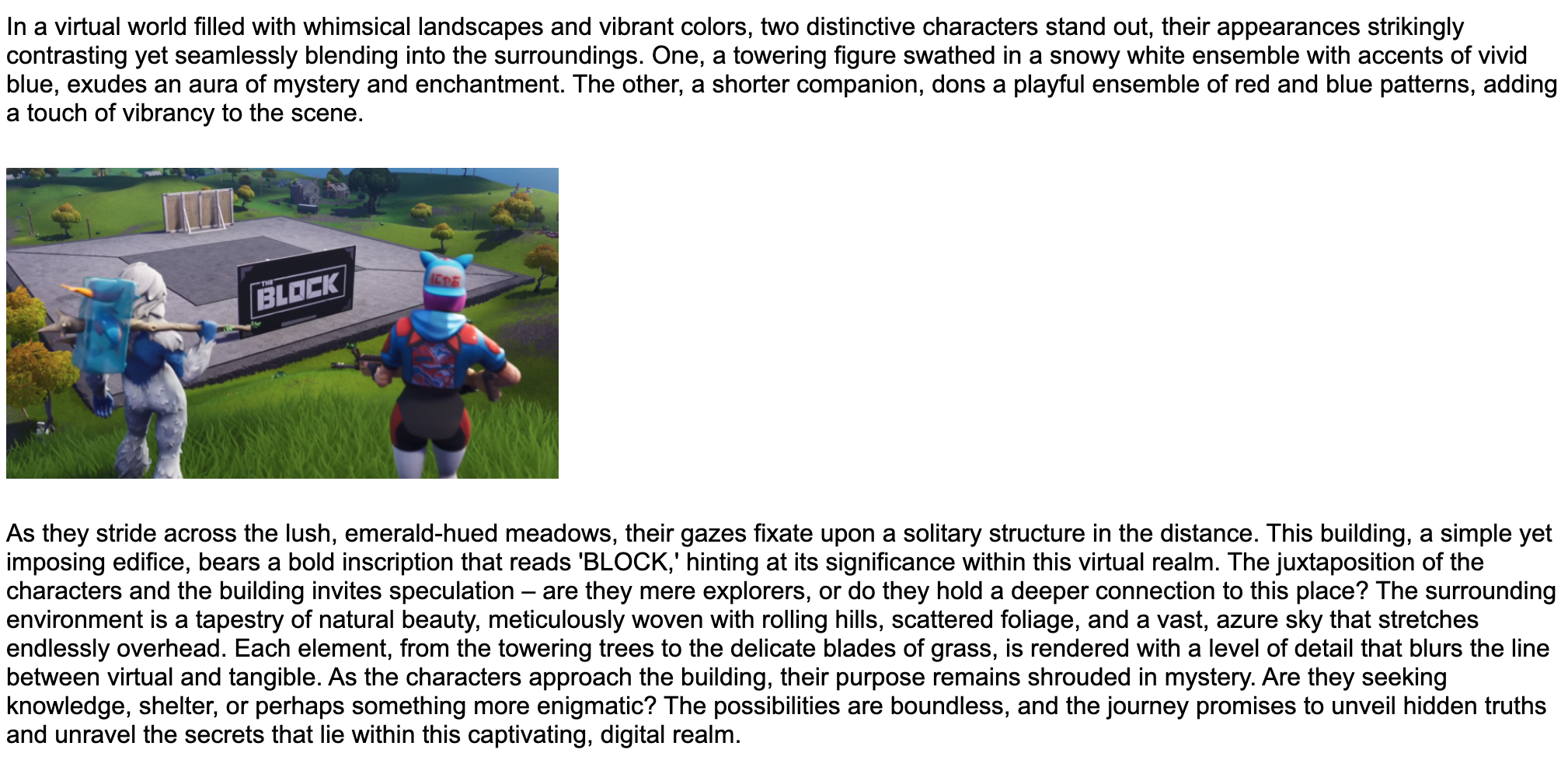} 
\caption{A positive synthetic document generated by Claude-3. The image is sampled from OBELICS dataset.
}
\label{fig:positive_doc}
\end{figure*}

\begin{figure*}[!ht] 
\centering 
\includegraphics[width=0.98\textwidth]{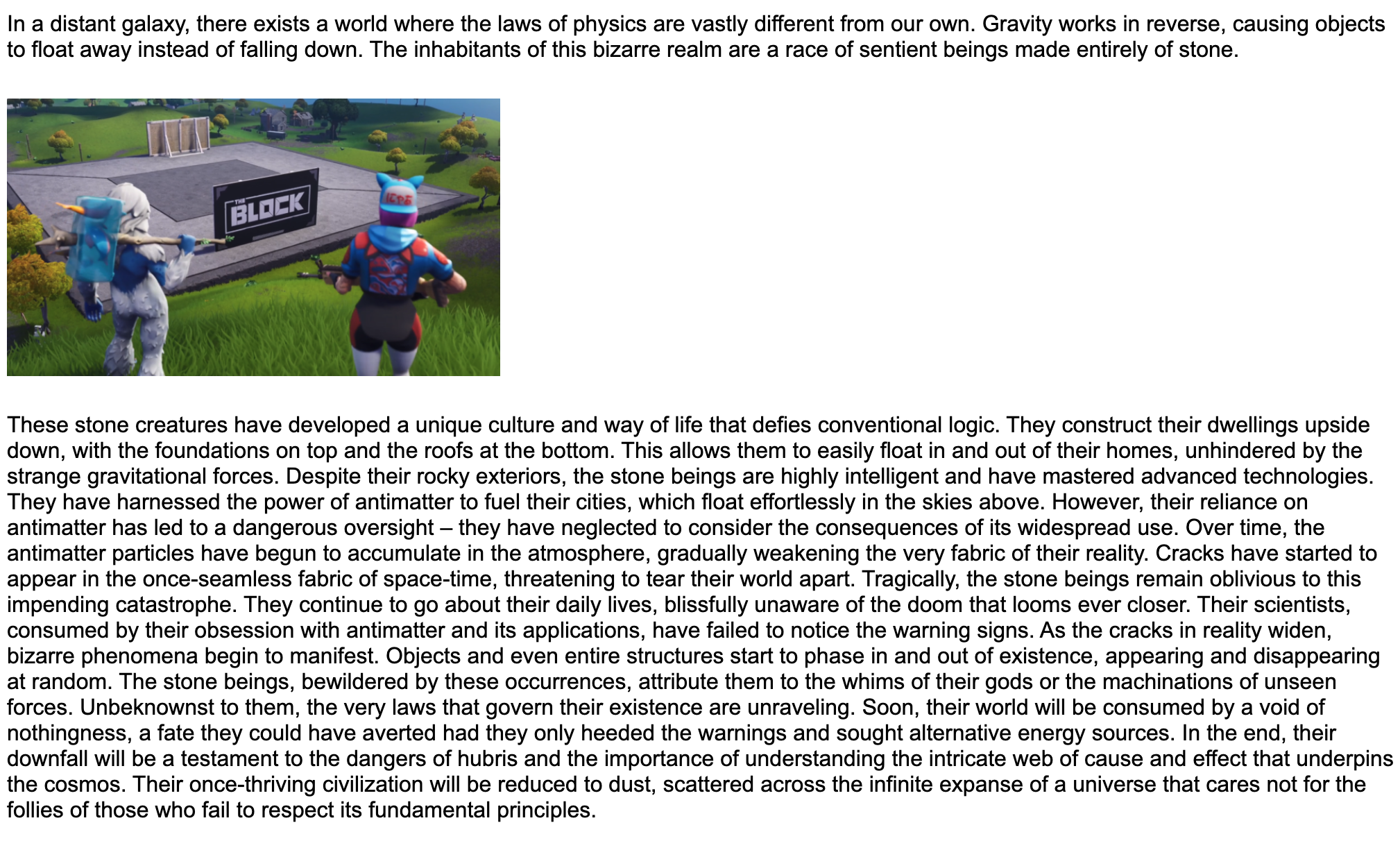} 
\caption{A hard negative synthetic document generated by Claude-3. The image is sampled from OBELICS dataset.
}
\label{fig:negative_doc}
\end{figure*}

We provide a pair of contrastive positive and hard negative synthetic documents in Figure~\ref{fig:positive_doc} and Figure~\ref{fig:negative_doc}. The positive document is apparently knowledge-intensive and has better detailed image-text alignment compared with the hard negative document.

\end{document}